\documentclass[conference]{IEEEtran}
\IEEEoverridecommandlockouts
\usepackage{cite}
\usepackage{amsmath,amssymb,amsfonts}
\usepackage{textcomp}
\usepackage{xcolor}
\def\BibTeX{{\rm B\kern-.05em{\sc i\kern-.025em b}\kern-.08em
    T\kern-.1667em\lower.7ex\hbox{E}\kern-.125emX}}

\long\def\comment#1{}
\usepackage{calc} 
\usepackage{enumitem} 

\usepackage{parskip}

\usepackage{algorithm}
\usepackage{algpseudocode}

\usepackage{times}
\usepackage{soul}
\usepackage{url}

\usepackage{setspace}
\usepackage{cite}
\usepackage{amsmath,amssymb,amsfonts}
\usepackage{graphicx}
\usepackage{textcomp}
\usepackage{xcolor}
\usepackage[detect-all]{siunitx}

\usepackage{xspace}
\usepackage{subcaption}
\usepackage{bm}
\usepackage{array}
\newcolumntype{P}[1]{>{\centering\arraybackslash}p{#1}}
\newcolumntype{M}[1]{>{\centering\arraybackslash}m{#1}}

\usepackage{algorithm,tabularx}
\usepackage{algpseudocode}
\makeatletter
\newcommand{\multilines}[1]{%
	\begin{tabularx}{\dimexpr\linewidth-\ALG@thistlm}[t]{@{}X@{}}
		#1
	\end{tabularx}
}
\makeatother

\usepackage{mathtools}

\DeclareMathOperator{\myspan}{span}

\usepackage{amsthm}
\usepackage{multirow}
\usepackage{color}
\usepackage{epstopdf}
\usepackage{endnotes}
\usepackage{framed}
\usepackage{cool} 
\usepackage{enumerate} 
	\usepackage[subnum]{cases}
	\usepackage{multicol}
	\usepackage{tablefootnote}
	
	\DeclareMathOperator*{\argmin}{arg\,min}

	\DeclarePairedDelimiterX{\norm}[1]{\lVert}{\rVert}{#1}
	\DeclarePairedDelimiterX{\abs}[1]{\lvert}{\rvert}{#1}
	\DeclarePairedDelimiterX{\innProd}[1]{\langle}{\rangle}{#1}
	\DeclarePairedDelimiter\set{\lbrace}{\rbrace}

	\newcommand{\SumNoLim}[2]{\ensuremath{\sum\nolimits_{#1}^{#2}}}
	\newcommand{\SumLim}[2]{\ensuremath{\sum\limits_{#1}^{#2}}}

	\newcommand{\BigP}[1]{\ensuremath{\Bigl(#1\Bigr)}}
	
	\newcommand{\BigC}[1]{\ensuremath{\Bigl\{#1\Bigr\}}}

	\theoremstyle{plain}

	\theoremstyle{definition}

\newcommand\blfootnote[1]{%
  \begingroup
  \renewcommand\thefootnote{}\footnote{#1}%
  \addtocounter{footnote}{-1}%
  \endgroup
}

\begin{document}
\bstctlcite{IEEEexample:BSTcontrol}

\title{Federated PCA on Grassmann Manifold for Anomaly Detection in IoT Networks\\

}

\author{\IEEEauthorblockN{Tung-Anh Nguyen, Jiayu He, Long Tan Le, Wei Bao, Nguyen H. Tran}
\IEEEauthorblockA{\textit{School of Computer Science} \\
\textit{The University of Sydney}\\
Sydney, Australia\\
\{tung6100, jihe5893\}@uni.sydney.edu.au, \{long.le, wei.bao, nguyen.tran\}@sydney.edu.au}
}

\maketitle

\begin{abstract}
In the era of Internet of Things (IoT), network-wide anomaly detection is a crucial part of monitoring IoT networks due to the inherent security vulnerabilities of most IoT devices. Principal Components Analysis (PCA) has been proposed to separate network traffics into two disjoint subspaces corresponding to normal and malicious behaviors for anomaly detection. However, the privacy concerns and limitations of devices' computing resources compromise the practical effectiveness of PCA. We propose a federated PCA learning using  Grassmann manifold optimization, which coordinates IoT devices to aggregate a joint profile of normal network behaviors for anomaly detection. First, we introduce a privacy-preserving federated PCA framework to simultaneously capture the profile of various IoT devices' traffic. Then, we investigate the alternating direction method of multipliers gradient-based learning on the Grassmann manifold to guarantee fast training and low detecting latency with limited computational resources. Finally, we show that the computational complexity of the Grassmann manifold-based algorithm is satisfactory for hardware-constrained IoT devices. Empirical results on the NSL-KDD dataset demonstrate that our method outperforms baseline approaches. 


\end{abstract}

\section{Introduction}
\blfootnote{This research is supported by SOAR funding.}
The realization of the Internet of Things (IoT) allows smart devices to share information and coordinate decisions over the Internet with little human intervention, which plays a remarkable role in improving the quality of our lives. In recent years, the number of IoT devices has increased unprecedentedly. However, as more IoT applications are applied across different sections, e.g., climate system~\cite{Afroz2018}, smart city~\cite{Zanella2014}, energy~\cite{Yang2016}, the detection of network attacks becomes an increasingly paramount task. In fact, many of the IoT devices have inherent security vulnerabilities due to limited computing resources~\cite{Cook2020}. 

To cope with this problem, an intrusion detection system (IDS) is used to detect possible attacks and anomaly~\cite{Casas2012,Nguyen2019, Wang2021}. Within an IDS deployment, local IoT devices typically connect to a local access gateway, which sends the real-time local information to the global security gateway. The global gateway retrieves the corresponding anomaly detection model~\cite{Javaid2016} to local gateways for intrusion detection. Subsequently, the IDS continuously monitors the communications of IoT devices and detects abnormal communication behavior using anomaly detection models.

The main objective of the anomaly detection model is to profile the normal behaviors and detect anomalies as deviations from the learned profile. Different models have been proposed to solve such a problem. Anomaly detection based on Principle Component Analysis (PCA) has been applied to successfully detect a wide variety of anomalies~\cite{Pascoal2012,Yu2017}. Given data observed from devices, PCA seeks to abstract the key relationships representation between these values, and then identify irregularities of new observations by their effect on the extracted representation. Although PCA can capture hidden states within the data and require a short detection time compared to \emph{supervised learning methods}~\cite{Cook2020}, it is impractical to upload the entire data collected by the IoT network due to the computational cost. Also, privacy is an important requirement for the IoT, where management and monitoring of the IoT should take place without sharing the users' data~\cite{Al-Fuqaha2015}. 

Recently, a Federated Learning (FL)-based IDS has been proposed to solve the privacy problem of anomaly detection~\cite{Nguyen2019,Mothukuri2022,Belenguer2022}. In federated IDS, multiple local gateways are treated as FL clients participating in the anomaly detection. During the training process, local gateways keep their data intact locally as clients, and only the learned models are transferred to the global security gateway. Most current works use the \emph{supervised learning methods} to detect intrusion, however, faces the challenges of producing labeled network datasets and detecting unknown anomaly upon which it has not been trained~\cite{Casas2012,Belenguer2022}. Furthermore, the training of a complex model is usually computationally expensive in terms of time and memory requirements for an IDS.

In this paper, we propose a novel FL framework for intrusion detection systems in IoT networks using PCA, an effective \emph{unsupervised anomaly detection} approach, to address above challenges. To build up an intrusion detection system on the FL framework, we first formulate a novel federated PCA optimization problem, namely Fed-PCA, to learn the profile of normal behavior in distributed network datasets. We then design an alternating direction method of multipliers (ADMM)-based algorithm FedPE to solve the proposed Fed-PCA problem. Specifically, in FedPE, each local gateway aims to find a local low-rank representation matrix on Euclidean space to capture the most variability in its own data such that the reconstruction error, averaged over all clients, is minimized in a distributed setting. The optimal representation matrix is treated as the hidden profile of normal behaviors of the IoT network, which deviates significantly in anomalous observations. 
While most conventional FL works do not consider the constraint, our Fed-PCA contains an orthogonal matrix constraint which lies on manifold. Therefore, we make use of optimization technique for the next proposed Fed-PCA on manifold algorithm. 

Due to a special manifold structure of the constraint in Fed-PCA problem, we further study the Grassmann manifold~\cite{zhang2018grassmannian} and propose an ADMM-based Grassmannian learning algorithm, called FedPG.  Note that the Grassmann manifold has been well-studied in exploiting the low-dimensional subspace which is tailor-made for manifold-valued data like network traffic~\cite{Patwari2005}. Therefore, the strength of FedPG lies in its capability to harness the structural information embedded in the problem, leading to faster convergence and better anomaly detection performance.
To show the effectiveness of our proposed methods, we compare the intrusion detection results with baseline approaches over the NSL-KDD dataset~\cite{Tavallaee2009}.

The main contributions of this work can be summarized as follows:
\begin{itemize}
    \item We formulate a new privacy-preserving federated PCA optimization problem (Fed-PCA) for IDS-based anomaly detection. By adopting the ADMM method, we propose a federated algorithm on Euclidean space, namely FedPE, to solve the proposed Fed-PCA, thereby capturing the profile of benign network traffic on decentralized datasets.
    \item We propose a new algorithm designed for federated PCA on Grassmann manifold (FedPG). The FedPG has a key advantage: the parameters of FedPG lie on the Grassmann manifold created by the constraint of the optimization problem, which leads to low complexity and a high convergence rate.
    \item The experimental results over the NSL-KDD dataset~\cite{Tavallaee2009} show that our Fed-PCA outperforms baselines including self-learning PCA in terms of detection accuracy, and capability of detecting a novel anomaly. The extensive evaluation demonstrates the robustness of Fed-PCA to non-i.i.d. distribution data which is suitable for intrusion detection systems (IDS). 
\end{itemize}
\section{Related Work}
\subsection{Federated Learning} 
FL is an emerging distributed learning paradigm enabling thousands of participating devices to periodically train and update machine learning models while preventing data leakage~\cite{Konecny2015}. Specifically, in FL, a model is downloaded and locally trained on clients' devices using their local data. These local models are then returned from the devices to the central server for aggregation (e.g., averaging the weights~\cite{fedavg}). The nature of communication-efficient and privacy-preserving makes FL beneficial to IoT networks and applications. Since the inception of the de-facto FedAvg~\cite{fedavg}, many other FL variants have been introduced to boost the intelligence of IoT systems in a wide range of applications~\cite{fl_iot}.

\subsection{Network Anomaly Detection} 
As IoT systems are prone to vulnerabilities~\cite{Neshenko2019}, anomaly-based IDS plays a critical role in safeguarding networks against different malicious activities, as the revolutionary advances in machine learning. Most of the early works on anomaly-based IDS in IoT networks have focused on centralized approaches (\cite{Meidan2018, Anthi2019, Pajouh2019, Zolanvari2019, Wang2021, Qiu2021}) in which data is collected to a central server raising privacy concerns and cyber invasions.

Recently, employing FL frameworks for network anomaly detection tasks has been on the horizon. For instance, Nguyen et al.~\cite{Nguyen2019} proposed a federated intrusion detection system using Recurrent Neural Network models trained on local gateways, then aggregating the local models into a global model. Wang et al.~\cite{Wang2022} showed a feed-forward artificial neural network can be used to improve the detection performance as the base model. Mothukuri et al.~\cite{Mothukuri2022} leveraged federated training on Gated Recurrent Units (GRUs) models and keeps the data intact on local IoT devices by sharing only the learned weights with the central server of FL.  However, due to the inherent imbalance between the normal traffic and abnormal traffic in IoT networks, those supervised learning approaches have been challenging in precisely detecting unknown anomalous activities~\cite{Cook2020}.
\vspace{-1pt}
\subsection{Principle Component Analysis} 
Over the past decade, PCA~\cite{Jolliffe2014}, a powerful unsupervised learning model, has been widely adopted in centralized machine learning. From network security perspectives, PCA-based anomaly detection has emerged as an efficient method, where PCA is used to separate a given dataset into two subspaces representing benign and malicious behavior~\cite{Lakhina2004}. Considering the temporal correlation of the data, \cite{Brauckhoff2009} proposed the use of the Karhunen-Loeve expansion, while it only uses data from a single router. 
Kernel PCA is proposed to analyze features from both perspectives of variance and mean~\cite{Chen2021}. Although the above methods have achieved promising results in anomaly detection, they only use data from a single device instead a federation of devices; therefore they can not be directly used for IDS in a resource-limited computing and privacy-conscious setting.

The idea of distributed PCA has become prevalent in the machine learning community in recent years~\cite{Wu2018}. For instance, Ge et al.~\cite{pmlr-v84-ge18a} suggested a minimax optimization framework to generate a distributed sparse PCA estimator while keeping privacy preservation. Grammenos et al.~\cite{Grammenos2020} proposed an asynchronous FL approach to execute PCA in a memory-limited setting. Based on the assumption that clients are arranged in a tree structure, the author introduced an algorithm that incrementally computes local model updates using a streaming procedure and adaptively estimates its leading principal components. 

In this work, we take a different approach to design an efficient solution for network intrusion detection in IoT systems. Our approach relies on an unsupervised federated PCA learning framework which not only address the privacy concerns, but also perform anomaly detection in an efficient way to prevent both known and never-before-seen-attacks. 
 

\section{Problem Description and Background}
\subsection{System Model}
\begin{figure}[t]
	\centering    
	\includegraphics[width=0.9\columnwidth]{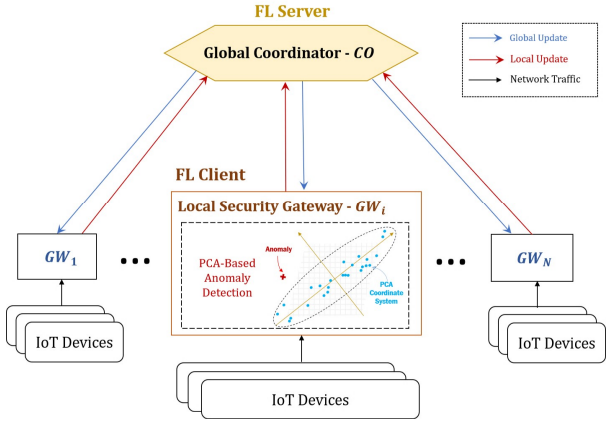}
	\caption{The visualization of the Federated PCA-based anomaly detection system model.}
	\label{fig:system}
\end{figure}
As shown in Fig~\ref{fig:system}, we consider a federated IDS for IoT networks. Typically, IoT systems contain multiple local gateways which each operates and monitors network traffic on a group of IoT devices. By integrating the FL processes into the gateways, we make them become security components (i.e. FL clients) performing the model training tasks on their local traffic. Those local security gateways, once deployed, will communicate with a global coordinator (i.e, FL server), which can be a cloud server, to achieve anomaly detection in a heterogeneous system. The advantage of federated IDS will not only help protect networks from malicious activities but also preserve data privacy, alleviate the computational burden of centralized approaches, improve bandwidth utilization, and further deal better with a surge of diverse data communications~\cite{Nguyen2019,Wang2022}.

The FL process between local security gateways ($GW$) and the global coordinator ($CO$) can be explained as follows. In the training phase, $CO$ and $GW$ will involve in a server-client computation to obtain the low-rank PCA representation of network traffic. Particularly, at each communication iteration, $CO$ broadcasts the current global model to $GW$s. After that, $GW$s update local models using their local data; $CO$ then aggregates the latest local models from a subset of sampled clients to update a new global model, and iterated until convergence. Once the training process is finished, $GW$s will leverage the final global model to perform PCA-based anomaly detection tasks in an unsupervised manner. We show in Sec.~\ref{Sec:FedPCA} that by using unlabelled data to train the model in a distributed setting, our proposed approach can help efficiently identify both known and unknown intrusions which are more generalizable than supervised learning approaches. Moreover, the proposed federated IDS system based on PCA is fast and lightweight (details in Sec.~\ref{Sec:FedPCA}) harmonizing with the limited-resource computing IoT devices.

\subsection{PCA for Centralized dataset}\label{Sec:CentralizedPCA}
Here, we introduce the PCA approach~\cite{Pearson1901} for centralized anomaly detection within the above system model. Assuming that each local gateway collects data, and sends the collected data to the global server, we use $X \in \mathbb{R}^{d \times D}$ to denote the fully observed dataset, where $d$ is data dimension and $D$ is the total number of data points. The dataset $X=[X_{1},X_{2}, \ldots, X_{N}]$ is distributed to $N$ clients. Each client $i \in \set{1,\ldots,N}$ contains a local dataset of $D_i$ records denoted by  $X_i = [x_{1},x_{2}, \ldots, x_{D_i}] \in \mathbb{R}^{d \times D_i}$, where $x_1\in \mathbb{R}^{d}$ represents a  record and $\SumNoLim{i=1}{N}D_i = D$.
Formally, PCA projects a given set of data onto principal components ordered by the amount of data variance that they capture~\cite{Jolliffe2002}. The PCA problem seeks the best rank-$k$ ($k < d$) matrix $\tilde{X}$ such that the reconstruction error, measured by $\norm{X - \tilde{X}}_F$, is minimized where $||\cdot||_F$ denotes the Frobenius norm~\cite{Udell2016}. Typically, the solution $\tilde{X}$ is based on the singular value decomposition of $X$, i.e.,
	$X = U \Sigma V^\top$,
where $U \in \mathbb{R}^{d \times d}$ contains the left singular vectors of $X$ on its columns, $\Sigma = \text{diag}(\sigma_1, \ldots, \sigma_d)$ contains the singular values ($\sigma_1 \geq \sigma_2 \geq \ldots \geq \sigma_d$), and $V \in \mathbb{R}^{D \times D}$ contains the right singular vectors of $X$ on its columns. Note that the columns of $U$ are orthonormal. Let $U_k \in \mathbb{R}^{d \times k}$ be the first $k$ columns of $U$ and $\Sigma_k = \text{diag}(\sigma_1, \ldots, \sigma_k)$. We can rewrite the closed-form solution for $\tilde{X}$ as
\begin{align} \label{E:pca_centralized}
	\tilde{X} = U_k U_k^\top X,
\end{align}
which can be interpreted as the projection of data matrix $X$ onto the column space of $U_k$. 
The PCA problem can then be represented as the following optimization problem:
\begin{equation} \label{Prob:pca_centralized}
	\begin{aligned}
		\min_{U \in \mathbb{R}^{d \times k}} \quad & \norm{(I - U U^\top)X}_F^2 \\
		\mathrm{s.t.} \quad & U^\top U = I.
	\end{aligned}
\end{equation}
The above problem aims to find an orthogonal matrix $U$, captured by its constraint, such that $UU^\top X$ is the best rank-$k$ approximation of $X$.

\subsection{PCA for Network Anomaly Detection}\label{Sec:PCA-for-AD}
As discussed above, PCA aims to minimize the reconstruction error, i.e. $\norm{X - \tilde{X}}_F$, to capture the most salient information of the original features in such a way that they can reconstruct the original feature set from the reduced feature set. Given an observed record $x\in \mathbb{R}^{d}$, its reconstruction error is denoted as $||x - U_k U_k^\top x||_F^2$. Considering the fact that anomaly is rare and presumably different than normal behaviors~\cite{Patel2019}, the gist of PCA-based anomaly detection is that the anomalous records should exhibit a larger reconstruction error. In practice, we use PCA to learn a profile that represents normal behavior. As in previous works~\cite{Brauckhoff2009}, an anomaly is flagged when the reconstruction error surpasses a security threshold.  

\section{Federated PCA for Network Anomaly Detection}\label{Sec:FedPCA}
\subsection{PCA for distributed dataset}
In the previous, we see that finding $U_k$ allows us to reduce the dimension of the dataset. The quantity $Z_k = U_k^\top X \in \mathbb{R}^{k \times D}$ in \eqref{E:pca_centralized} is the lower-dimensional version of $X$ such that the reconstruction error is minimum. In the distributed setting, we aim to learn a \emph{common representation} among all client's dataset. In other words, we aim to find the rank-$k$ matrix $U$ such that the reconstruction error, averaged over all clients, is minimized. Inspired by the finite-sum formulation, we introduce the federated PCA Problem as:
\begin{equation} \label{Prob:pca_decentralized}
	\begin{aligned}
		\min_{U \in \mathbb{R}^{d \times k}} \quad & \SumLim{i=1}{N} \BigC{f_i(U) := \norm{(I - U U^\top)X_i}_F^2} \\
		\mathrm{s.t.} \quad & U^\top U = I. 
	\end{aligned}
\end{equation}

\subsection{ADMM for Federated PCA optimization}~\label{Sec:FedPE}
\begin{algorithm}[t]
	\caption{Federated PCA on Euclidean space (FedPE)}
	\label{algo1}
	\begin{algorithmic}[1]
		\State Randomly initialize $Z^0$ and $U_i^0, \forall i.$
		\For{$k = 0, \ldots, T - 1$} \Comment{\textit{Global rounds}}
		\For {client $i = 1, \ldots, N$ in parallel} \Comment{\textit{Local rounds}} 
		\State $U_i^{k+1} = \underset{U_i}{\operatorname{\argmin}}  \BigC{f_i(U_i) + \innProd{Y_i^k, U_i - Z^k}_F + \innProd{T_i^k, h_i(U_i)}_F + \frac{\rho}{2} \norm{U_i - Z^k}_F^2 + \frac{\rho}{2} \norm{h_i(U_i)}_F^2}$
		\EndFor
		\State  $CO$ updates $Z^{k+1} = \frac{1}{N} \SumNoLim{i=1}{N} U_i^{k+1}$  \Comment{\textit{Global update}}
		\State $CO$ broadcasts $Z^{k+1}$ to all clients
		\For {client $i = 1, \ldots, N$ in parallel} \Comment{\textit{Local update}}
		\State   $Y_i^{k+1} = Y_i^k + \rho \BigP{U_i^{k+1} - Z^{k+1}} $ 
		\State   $T_i^{k+1} = T_i^k + \rho \, h_i(U_i^{k+1})$ 
		\EndFor
		\EndFor
	\end{algorithmic}
\end{algorithm}
As discussed in Sec.~\ref{Sec:CentralizedPCA}, the centralized PCA achieves incorporation of local gateway for anomaly detection by sending their local data to a global server. However, within an IDS development, it is impractical to share local data with the server for communication costs and privacy concerns.
To address the problem~(\ref{Prob:pca_decentralized}) in a privacy-preserving manner, we proposed an algorithm on Euclidean space, namely Federated PCA on Euclidean space (FedPE). In order to design FedPE following an FL style, we need to transform \eqref{Prob:pca_decentralized} into a problem that enables distributed algorithm structure. 


By first introducing a set of local variables $U_1, \ldots, U_N$,  we have an equivalent problem to  \eqref{Prob:pca_decentralized} as follows:
\begin{equation} \label{Prob:pca_decentralized_admm}
	\begin{aligned}
		\min_{U_1, \ldots, U_N} \quad & { \SumLim{i=1}{N} \norm{(I - U_i U_i^\top)X_i}_F^2} \\
		\mathrm{s.t.} \quad & U_i = Z, \; \forall i=1,\ldots,N. \\
		 \quad & U_i^\top U_i = I, \; \forall i=1,\ldots,N, 
	\end{aligned}
\end{equation}
where the first linear constraint enforces the ``consensus'' \cite{Boyd2011} for all local $U_i$. The second constraint, however, is not linear in $U_i$. Using the technique in \cite{giesen2016distributed} to handle non-linear constraints in ADMM, we next define the following function 
\begin{align*}
	h_i (U_i) = \max{\set{0, U_i^\top U_i - I}}^2,
\end{align*}
where the operators $\max$ and squared are component-wise. Then, we have the problem \eqref{Prob:pca_decentralized_admm} equivalent to the following
\begin{equation} \label{Prob:pca_decentralized_admm_nonlinear}
	\begin{aligned}
		\min_{U_1, \ldots, U_N} \quad & { \SumLim{i=1}{N} \norm{(I - U_i U_i^\top)X_i}_F^2} \\
		\mathrm{s.t.} \quad & U_i = Z, \; \forall i=1,\ldots,N \\
		 \quad & h_i(U_i) = 0, \; \forall i=1,\ldots,N.
	\end{aligned}
\end{equation}
The augmented Lagrangian function associated with~(\ref{Prob:pca_decentralized_admm_nonlinear}) is
\begin{align*}
	\mathcal{L}_{\rho}(U, Z, Y, T) &= \SumLim{i=1}{N} f_i(U_i) \\
	& + \SumLim{i=1}{N}\innProd{Y_i, U_i - Z}_F + \SumLim{i=1}{N}\innProd{T_i, h_i(U_i)}_F  \\
	& + \frac{\rho}{2} \SumLim{i=1}{N} \norm{U_i - Z}_F^2 + \frac{\rho}{2} \SumLim{i=1}{N} \norm{h_i(U_i)}_F^2,
\end{align*}
where $\innProd{\cdot, \cdot}_F$ is the Frobenius inner product between two same-sized matrices where $\innProd{A, B}_F = \sum_{i, j} A_{i, j} B_{i, j} ~\text{and $A$ and $B$ are real matrices}$.

The gradient of the augmented Lagrangian with respect to dual variables $Y_i$ and $T_i$ are $\nabla_{Y_i} \mathcal{L}_{\rho}(U, Z, Y, T) = U_i - Z$ and $\nabla_{T_i} \mathcal{L}_{\rho}(U, Z, Y, T) = h_i(U_i)$.
Therefore, the ADMM updates with step side $\rho$ in each iteration $k$ \cite{Boyd2011} are
\begin{align}
	U_i^{k+1} &= \argmin_{U_i} \BigC{f_i(U_i) + \innProd{Y_i^k, U_i - Z^k}_F + \innProd{T_i^k, h_i(U_i)}_F \nonumber \\ 
		& \quad \quad \quad \quad \quad  + \frac{\rho}{2} \norm{U_i - Z^k}_F^2 + \frac{\rho}{2} \norm{h_i(U_i)}_F^2} \label{ADMM_PCA_update_U_i}. \\
	Z^{k+1} &= \frac{1}{N} \SumLim{i=1}{N} \BigP{U_i^{k+1} + \frac{1}{\rho} Y_i^k} \label{ADMM_PCA_update_Z}. \\
	Y_i^{k+1} &= Y_i^k + \rho \BigP{U_i^{k+1} - Z^{k+1}} \label{ADMM_PCA_update_Y_i}. \\
	T_i^{k+1} &= T_i^k + \rho \, h_i(U_i^{k+1}). \label{ADMM_PCA_update_T_i}
\end{align}
Substituting $Z^{k+1}$ to $Y_i^{k+1}$ leads to $\frac{1}{N} \SumNoLim{i=1}{N}Y_i^{k+1}=0$, which means $Z$ update can be rewritten as \cite{Boyd2011}:
\begin{align}
    Z^{k+1} &= \frac{1}{N} \sum_{i=1}^{N} U_i^{k+1}.
\end{align}
The whole FedPE scheme is presented in Algorithm~\ref{algo1}.
The convergence of ADMM-based algorithms has been well-studied in the literature. According to the analysis results in \cite{giesen2016distributed, Hong2015,Chang2016}, the proposed FedPE algorithm guarantees convergence to a set of stationary solutions.

\subsection{Grassmann Manifold}
\begin{figure}[t]
	\centering    
	\includegraphics[width=0.7\columnwidth]{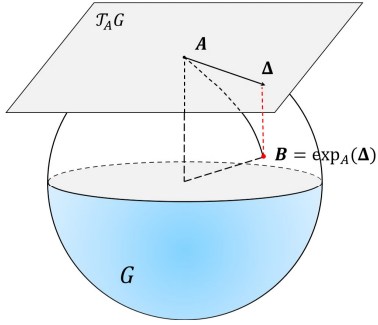}
	\caption{Example of moving on Grassmann manifold illustrated by a sphere. Given a point $A$ on Grassmann manifold $G$ and a vector $\Delta$ on the tangent space denoted by $\mathcal{T}_{A}G$ at $A$, a point $B$ is identified by exponential mapping $\exp_A(\Delta)$.  In our context, the manifold $G$ contains all matrix $U$ satisfy the condition $U^{\top}U = I$. }
	\label{fig:gm}
\end{figure}
While FedPE can efficiently solve the proposed optimization problems, we further generalize the Federated PCA on Grassmann manifold (FedPG) as a special case of FedPE for faster convergence. Specifically, the constraint introduced in~(\ref{Prob:pca_decentralized}) implies that the parameters of (\ref{ADMM_PCA_update_U_i}) can be embedded into Grassmann manifold subspace~\cite{Jiayao2018}. In the below discussion, we first introduce the notion of the Grassmann manifold and then correlates them to our proposed algorithm in Sec.~\ref{FedPG}.

Given integers $n \geq k >0$, Grassmann manifold~\cite{Jiayao2018}, denoted as $G(n,k)$, is created by all $k$-dimensional subspaces of the Euclidean space $\mathbb{R}^n$, e.g., the surface of a sphere in a 3D space depicted in Fig.~\ref{fig:gm}. According to the definition of Grassmann manifold, every element in $G(n,k)$ can be identified by selecting $n \times k$ orthogonal matrix $A$ spanning the corresponding subspace.
\begin{equation}\label{eq:grass-constraint}
	{G}(n, k) = \{ \myspan(A): A \in \mathbb{R}^{n\times k}, A^\top A= I_k\}.
\end{equation}
In the Grassmann manifold, the notion of tangent space plays a key role. Given a point $A\in G(n,k)$, a tangent space at $A$ is defined as a set of vectors $\Delta$ such that $A^{\top}\Delta = 0$.  Intuitively, a tangent space is a  vector space that contains all possible directions in which one can tangentially pass through $A$.

Consider a function $F: {G(n,k) \rightarrow \mathbb{R}}$ and its Euclidean gradient $F_A$ at point $A$, 
the projected gradient of the function $F$ on the Grassmann manifold is computed by projecting the Euclidean gradient $F_A$ onto the tangent space of the Grassmann manifold using the orthogonal projection $F_A \rightarrow \nabla_A{F}$:
\begin{equation}
	\nabla_AF\ = (I_k - AA^{\top})F_A.
\end{equation}
The aim is to compute a forward step in the direction of the projected gradient on the Grassman manifold. While the projection from the tangent plane to the manifold can be done by using exponential mapping, this approach is expensive in terms of computation. Alternatively, the retraction function is introduced as a local approximation of exponential mapping on the manifold with less computation complexity \cite{sato2019riemannian}. 
Thus, a retraction operation is performed by mapping the tangent matrix back on the manifold:
\begin{equation}
	B = R(A -\eta \nabla_A F),
\end{equation}
where $\eta$ is the step size and $R$ is the retraction function on the Grassmann manifold performed by applying QR decomposition \cite{Jiayao2018}. 

\subsection{Distributed PCA using ADMM and Grassmann manifold}~\label{FedPG}
\begin{algorithm}[t]
	\caption{Federated PCA on Grassmann manifold (FedPG)}
    \label{algo2}
	\begin{algorithmic}[1]
		\State Randomly initialize $Z^0$ and $U_i^0, \forall i$
		\For{$k = 0, \ldots, T - 1$} \Comment{\textit{Global  rounds}}

		\For {client $i = 1, \ldots, N$ in parallel} 
 		\State $U^{k, 0}_i = U^{k}_i$
		\For{$c = 0, \ldots, C - 1$} \Comment{\textit{Local rounds}}
		\State $\nabla_U F_i^k(U_i^{k,c}) = (I_d - U_i^{k,c}U_i^{k,c \top}) F_i^k(U_i^{k,c})$ 
		\State $U_{i}^{k, c+1} =R(U_i^{k, c} - \eta \nabla_U F_i^k(U_i^{k,c}))$
		\EndFor
		\State $U_i^{k+1} = U_{i}^{k, C} $
		\EndFor
		\State  $CO$ updates $Z^{k+1} = \frac{1}{N} \SumNoLim{i=1}{N} U_i^{k+1}$\Comment{\textit{Global update}}
		\State $CO$ broadcasts $Z^{k+1}$ to all clients 
		\For {client $i = 1, \ldots, N$ in parallel} \Comment{\textit{Local update}}
		\State   $Y_i^{k+1} = Y_i^k + \rho \BigP{U_i^{k+1} - Z^{k+1}} $ 
		\State   $T_i^{k+1} = T_i^k + \rho \, h_i(U_i^{k+1})$ 
		\EndFor
		\EndFor
	\end{algorithmic}
\end{algorithm}
Recapping the introduced problem~\eqref{Prob:pca_decentralized_admm_nonlinear}, we next propose an algorithm, namely Federated PCA on Grassmann manifold (FedPG), summarized in Algorithm~\ref{algo2}. Similar to the previous solution in Sec~\ref{Sec:FedPE}, we first enforce the consensus constraints through ADMM, then the augmented Lagrangian is
\begin{equation}
\begin{split}
	\mathcal{L}(U, Z, Y) &= \SumLim{i=1}{N} f_i(U_i) + \SumLim{i=1}{N} Y_i^\top (U_i - Z) \\
	&+ \frac{\rho}{2} \SumLim{i=1}{N} \norm{U_i - Z}^2_F.
\end{split}
\end{equation}
We re-form the update rules for $U_i$ and $Z$ on the Grassmann manifold as follows: 
\begin{align} \label{E:Fopt}
	U_i^{k+1} &= \argmin_{U_i} \BigC{F_i^k(U_i)} \quad \mathrm{s.t.}
\end{align}

\begin{align} \label{E:Fopt_2}
	F_i^k(U_i) &:= f_i(U_i) + Y_i^{k\top}(U_i - Z^k) + \frac{\rho}{2} \norm{U_i - Z^k}_F^2.
\end{align}
In Algorithm \ref{algo2}, as shown in lines 5-8, projected gradient descent on the Grassmann manifold is applied at every local iteration in client for finding the update of $U_i^{k+1}$ in \eqref{E:Fopt}. Considering the definition of the Grassmann manifold in (11), we enforce the second constraint $U_i^TU_i = I$ by using the projected gradient descent on the Grassmann manifold (line 7) as follows
\begin{align}
	U^{k+1}_i = R(U^k_i - \eta \nabla_U F_i^k(U^k_i)), \label{E:Gpro1}
\end{align}
where $\eta$ is the step size, and $\nabla_U F_i^k(U^k_i)$ is the gradient of $F_i^k(U^k_i)$ at point $U_i^k$, and the retraction function $R(\cdot)$ is the projection operation. Pictorially, $U^{k}_i$ and $U^{k+1}_i$ are illustrated by points $A$ and $B$ in Fig.~\ref{fig:gm}, respectively. We compute the updates by projecting the Euclidean gradient onto the tangent space of the manifold using orthogonal projection $F_i^k(U^k_i)\rightarrow  \nabla_U F_i^k(U^k_i)$ (line 6 ) as follows.
\begin{equation}
	\nabla_U F_i^k(U^k_i) = (I_d - U_i^{k}U_i^{k \top})F_i^k(U^k_i). \label{E:Gpro2}
\end{equation}
Note that the two equations \eqref{E:Gpro1} and \eqref{E:Gpro2} are repeatedly updated in $C$ local rounds at lines 5-8 in Algorithm~\ref{algo2} to solve the equation \eqref{E:Fopt}. 

The updates of $Y_i^k, T_i^k$ and $Z_i^k$ are identical to (8), (9) and (10), respectively.


\subsection{Computational Complexity}\label{Sec:Complexity}
Here, we study the strength of computational complexity of the proposed algorithm FedPG. We recall that   $d$ and $k$ denote the input data dimension and the number of PCA components, respectively. $C$ and $T$ denote the number of local and global rounds, respectively.

In FedPG, the major local-client computational costs in each global communication round come from two parts, calculating the gradient updates for the matrix $U_i^k$ (Lines 7-8 in Algorithm~\ref{algo2}) and performing consensus updates for the matrices $Y$ and $T$ (Lines 15-16 in Algorithm~\ref{algo2}). Especially, computing the orthogonal projection to obtain the gradient costs at most $O(2d^2k + 2dk)$ flops for two matrix-matrix multiplications (Line 7 in Algorithm~\ref{algo2}). Once the gradient is calculated, the retraction of the update step on the Grassmann manifold is done by applying QR decomposition with complexity $O(dk + dk^2)$ (Line 8 in Algorithm~\ref{algo2}). The $Y$ and $T$ update phase needs simple matrix-matrix multiplication requiring $O(d^2k)$ flops (Line 16 in Algorithm~\ref{algo2}). Suppose the algorithm is terminated after $C$ rounds, then the overall computational complexity for each local security gateway is given by $O(C(d^2k + dk + dk^2))$. Therefore, FedPG's  complexity is (i) independent of the data side $D_i$, and (ii) dominated by two factors $d$ and $k$. In IDS applications, the network data is often not too high-dimensional (i.e, the value of $d$ is not too large~\cite{Thakkar2020} and $k < d$, as confirmed in the experiments in next section) making this computational complexity acceptable for limited-resource computing IoT devices.

At each global communication round, the global coordinator mainly updates the matrix $Z$ by averaging the sum of $U$ matrices from a subset $S = |10\%N|$ client number, hence costing $O(Sdk)$ flops.  This leads to the overall computational complexity of the global coordinator for $T$  rounds as $O(TSdk)$ flops.
\section{Experiment}
In this section, we first introduce the dataset, parameters setting, and performance evaluation metrics. In order to show the effectiveness of the proposed methods, we then compare Fed-PCA with several baseline approaches in an IDS deployment.

\begin{table*}[t]
\caption{NSL-KDD dataset attacks label taxonomy.}
\label{tab:NLS-KDD-label}
\centering
\begin{tabular}{lcc}
\hline
\textbf{Main Class} & \textbf{Sub-class Attacks in Training Set}                                                                              & \textbf{New Sub-class Attacks in Testing Set}                                                                        \\ \hline
Dos                 & back, land, neptune, pod, smurf, teardrop                                                                               & apache2, mailbomb, processtable, udpstorm                                                                            \\
Probe               & ipsweep, nmap, portsweep, satan                                                                                         & mscan, saint                                                                                                         \\
U2R                 & bufferoverflow, loadmodule, perl, rootkit                                                                               & ps, sqlattack, xterm                                                                                                 \\
R2L                 & \begin{tabular}[c]{@{}c@{}}ftp write, guesspasswd, imap, multihop, phf,  spy, \\ warezclient, warezmaster,\end{tabular} & \begin{tabular}[c]{@{}c@{}}httptunnel, named, sendmail, snmpgetattack, snmpguess,\\ xlock, xsnoop, worm\end{tabular} \\ \hline
\end{tabular}
\end{table*}

\begin{table}[t]
	\centering
	\caption{NSL-KDD Dataset}
	\label{tab:tab_3}
	\begin{tabular}{|c|c|c|c|c|}
		\hline
		\multirow{2}{*}{\textbf{Category}} & \multicolumn{2}{c|}{\textbf{Training Set}} & \multicolumn{2}{c|}{\textbf{Test Set}}  \\
		\cline{2-5}
		& \#Record & Rate(\%) & \#Record & Rate(\%) \\
		\hline
		\hline
		Normal & 67,343 & 53.46 & 9,711 & 43.08 \\
		\hline
		DoS & 45,927 & 36.46 & 7,458 & 33.08 \\
		\hline
		Probe & 11,656 & 9.25 & 2,421 & 10.74 \\
		\hline
		R2L & 995 & 0.79 & 2,754 & 12.22 \\
		\hline
		U2R & 52 & 0.04 & 200 & 0.89 \\
		\hline
		\hline
		\textit{Total} & \textit{125,973} & \textit{100} & \textit{22,544} & \textit{100} \\
		\hline
	\end{tabular}
\end{table}

\begin{table}[t]
	\centering
	\caption{Detection Performance (\%) over NSL-KDD.}
	\label{tab:performance}
	\begin{tabular}{|m{2.7cm}|m{0.7cm}|m{0.7cm}|m{0.7cm}|m{0.7cm}|m{0.7cm}|}
		\hline
		\textbf{Method}            & \textbf{Acc} & \textbf{Pre} & \textbf{TPR} & \textbf{FPR} & \textbf{F1} \\ 
		\hline
		\hline
		Fed-PCA (ours)    & \textbf{84.84} & \textbf{91.76} & \textbf{80.60} & \textbf{9.55}     & \textbf{85.82}  \\ 
		\hline
		Centralized PCA   & 84.79  & 91.73  & 80.57  & 9.59  &  85.79  \\ 
		\hline
		Self-learning PCA & 60.72  & 67.64  & 59.41  & 37.55 &  63.26  \\
		\hline
		Auto-Encoder      & 80.5   & 87.42  & 76.79  & 14.59 & 81.76  \\
		\hline
		LSTM              & 80.7   & 87.62  & 76.96  & 14.36 &  81.94  \\
		\hline
	\end{tabular}
\end{table}

\subsection{Experiment settings}

Our implementation for the proposed algorithms is PyTorch-based. The experiments are conducted on an Intel(R) Xeon(R) CPU @ 2.20GHz server with Ubuntu 18.04.5 LTS, 13GB RAM, and a NVIDIA Tesla P100 GPU.
\\
\subsubsection{Dataset and metrics}\label{Sec:dataset}
We conduct experiments with the widely used network attack dataset, NSL-KDD~\cite{Tavallaee2009}. Each NSL-KDD record contains connection and traffic features (e.g. connection duration,  protocol type, number of bytes, etc.), which are labeled either as normal traffic or one of the four classes of intrusions, i.e., Denial of service attacks (DoS), Probing attacks (Prob), unauthorized access to super-user privileges (U2R), and unauthorized access from a remote machine (R2L). The whole dataset consists of a training set (\textit{KDDTrain+}) with 125,973 records, and a testing set (\textit{KDDTest+}) with 22,544 records. Table~\ref{tab:tab_3} describes the detailed statistics of training and testing records for normal and different attack classes. As shown in Table~\ref{tab:NLS-KDD-label}, the NSL-KDD dataset contains a wide variety of sub-classes intrusions. The testing set contains $17$ new attack types not present in the training set, which will permit us to evaluate the effectiveness of FedPE and FedPG in detecting unknown attacks. 

In our experiments, we select a set of 34 continuous-valued features for training. Considering the fact that non-compromised devices usually generate only legitimate communications when first released~\cite{Nguyen2019,Liu2021}, we train our models with 67,343 normal records to capture the profile of normal traffic. We use five metrics in evaluation, i.e., Accuracy (Acc), Precision (Pre), True Positive Rate, False Positive Rate (FPR) and F1-score (F1). For anomaly detection, TPR is a measure of the classifier correctly detecting malicious samples of all malicious records. FPR is a rate of the classifier incorrectly classifying benign samples as malicious of all benign records. Proper anomaly detection can maximize the TPR with a relatively low FPR to make the detection system effective and stable. Among the metrics, F1-score is also a complete metric as a blend of the Precision and TPR~\cite{Belenguer2022}. Note that for unsupervised methods, a reconstruction error is computed using the learned profile for each record. Anomalies are detected when their records deviate significantly from the error threshold. We choose the $p$-th percentile of the sorted reconstruction error of samples as the threshold. We then use a Receiver Operating Characteristics (ROC) curve to show the trade-off between TPR and FPR under different thresholds~\cite{Brauckhoff2009}.
\\
\subsubsection{Federated setting}  
Within an IDS deployment, local IoT devices typically connect to a local gateway as an access client, which locally collects data to train the anomaly detection model. Considering the natural diversity of traffic distribution of different clients, we collect data for clients by splitting the whole training set into 20 sub-sets, each corresponding to the obtained local data with non-i.i.d. distributions. Specifically, we select subsets of training data by grouping the records in the order of the number of bytes (i.e. 'dst\_bytes'). To evaluate the proposed methods in a realistic IDS development setting, we learn our federated PCA models based on the records of local clients without sharing data between clients. We use the whole testing set with unknown attacks for all methods. Before training, we normalize each feature with a z-score function using the standard deviation and mean calculated in the corresponding training set.

In the following experiments, we select $k=30$ and $N=20$, which are fine-tuned values by conducting a grid search to achieve the best test performance. The sensitivity of the proposed algorithm with respect to $N$ is provided in Sec.~\ref{Sec:results-grassmann}. At each communication round, we randomly sample $|S_t| = 10\%$ of total number of clients to participate in training. The number of global and local rounds are set to $T=1000, C=30$, respectively. Each task shares the same hyper-parameters setting.

\subsection{Experimental Results}
\subsubsection{Detection performance} 
\begin{figure*}[t]
    \centering
    \begin{subfigure}{0.3\textwidth}
      \centering
      \includegraphics[width=\textwidth]{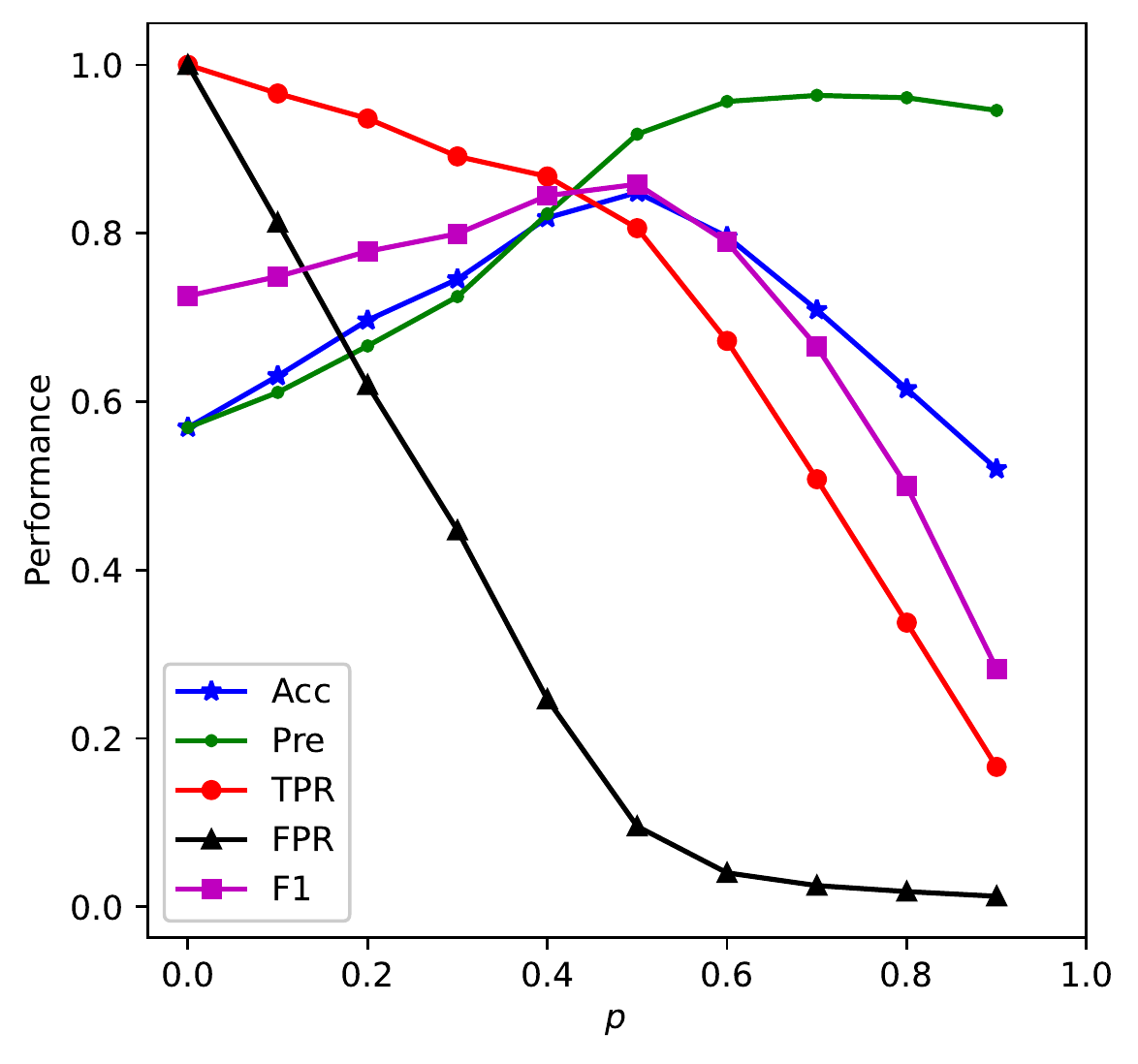}
      \caption{}
      \label{fig:results-p}
    \end{subfigure}%
    \begin{subfigure}{0.3\textwidth}
      \centering
      \includegraphics[width=\textwidth]{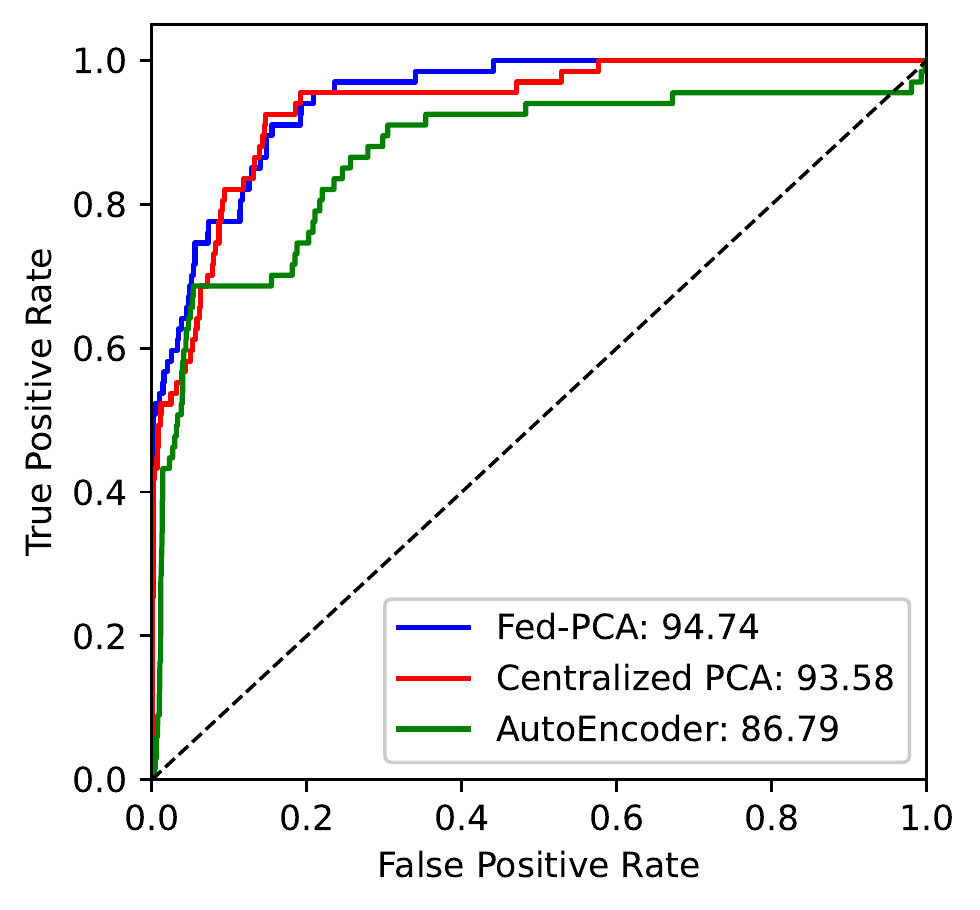}
      \caption{}
      \label{fig:roc_U2R}
    \end{subfigure}
    \begin{subfigure}{0.3\textwidth}
      \centering
      \includegraphics[width=\textwidth]{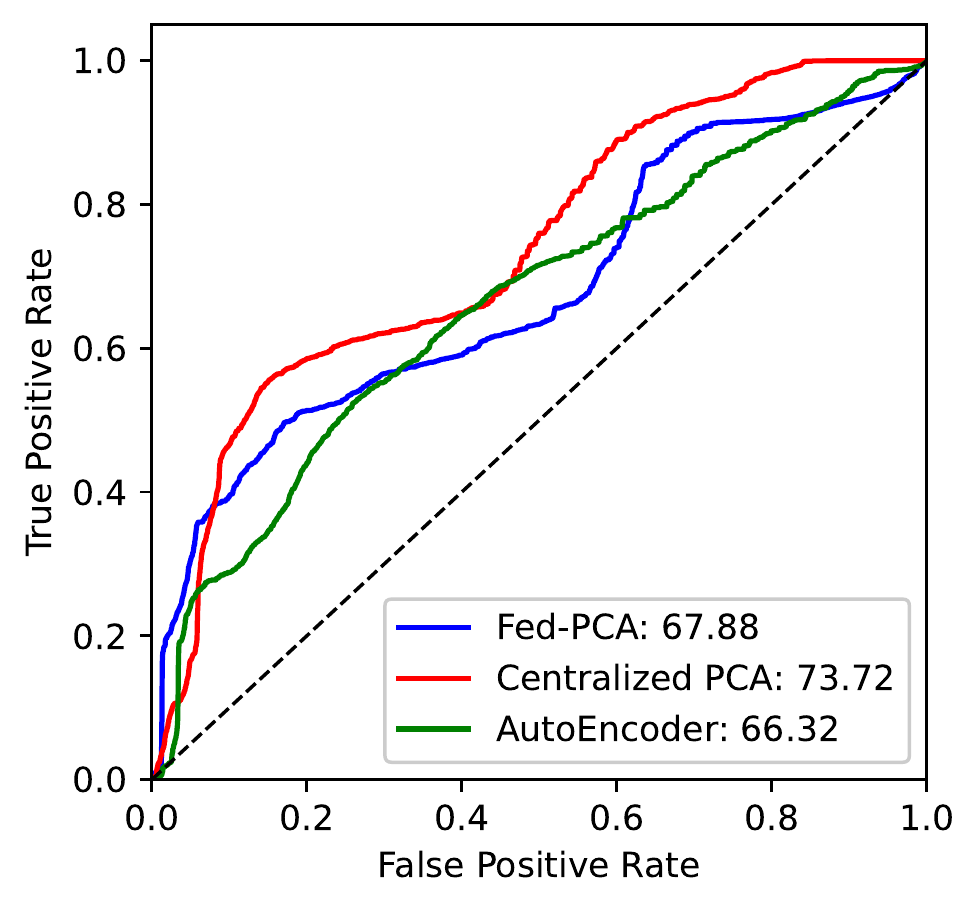}
      \caption{}
      \label{fig:roc_R2L}
    \end{subfigure}
  \caption {(a) Performance of Fed-PCA with different threshold $p$. (b) ROC curves over U2R attacks. (c) ROC curves over R2L attacks. The AUC scores are shown in the legends.}
  \label{fig:abcd}
\end{figure*}

\begin{figure*}[t]
    \centering
    \begin{subfigure}{0.3\textwidth}
      \centering
      \includegraphics[width=\textwidth]{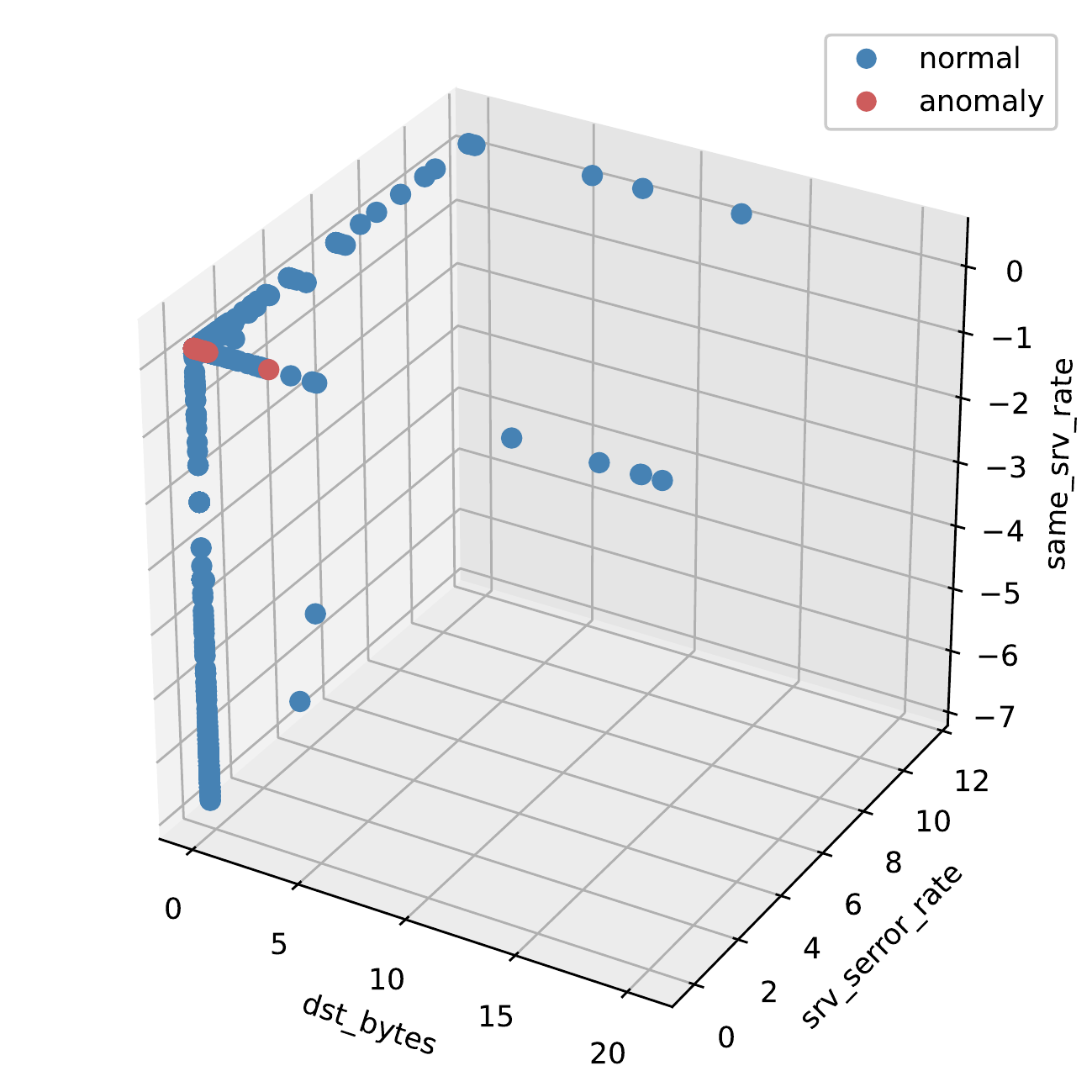}
      \caption{}
      \label{fig:sc-original}
    \end{subfigure}%
    \begin{subfigure}{0.3\textwidth}
      \centering
      \includegraphics[width=\textwidth]{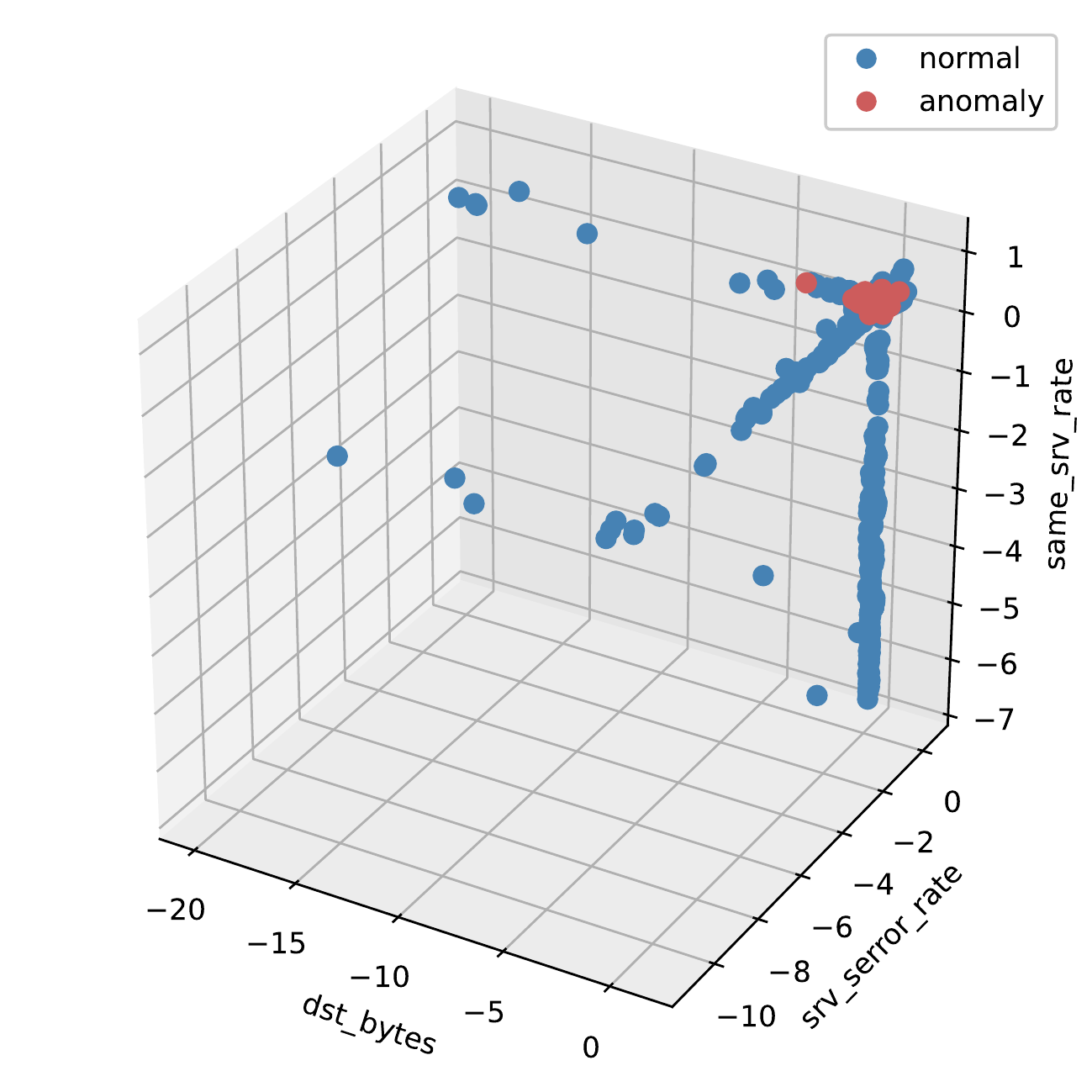}
      \caption{}
      \label{fig:sc-c-transformed}
    \end{subfigure}
    \begin{subfigure}{0.3\textwidth}
      \centering
      \includegraphics[width=\textwidth]{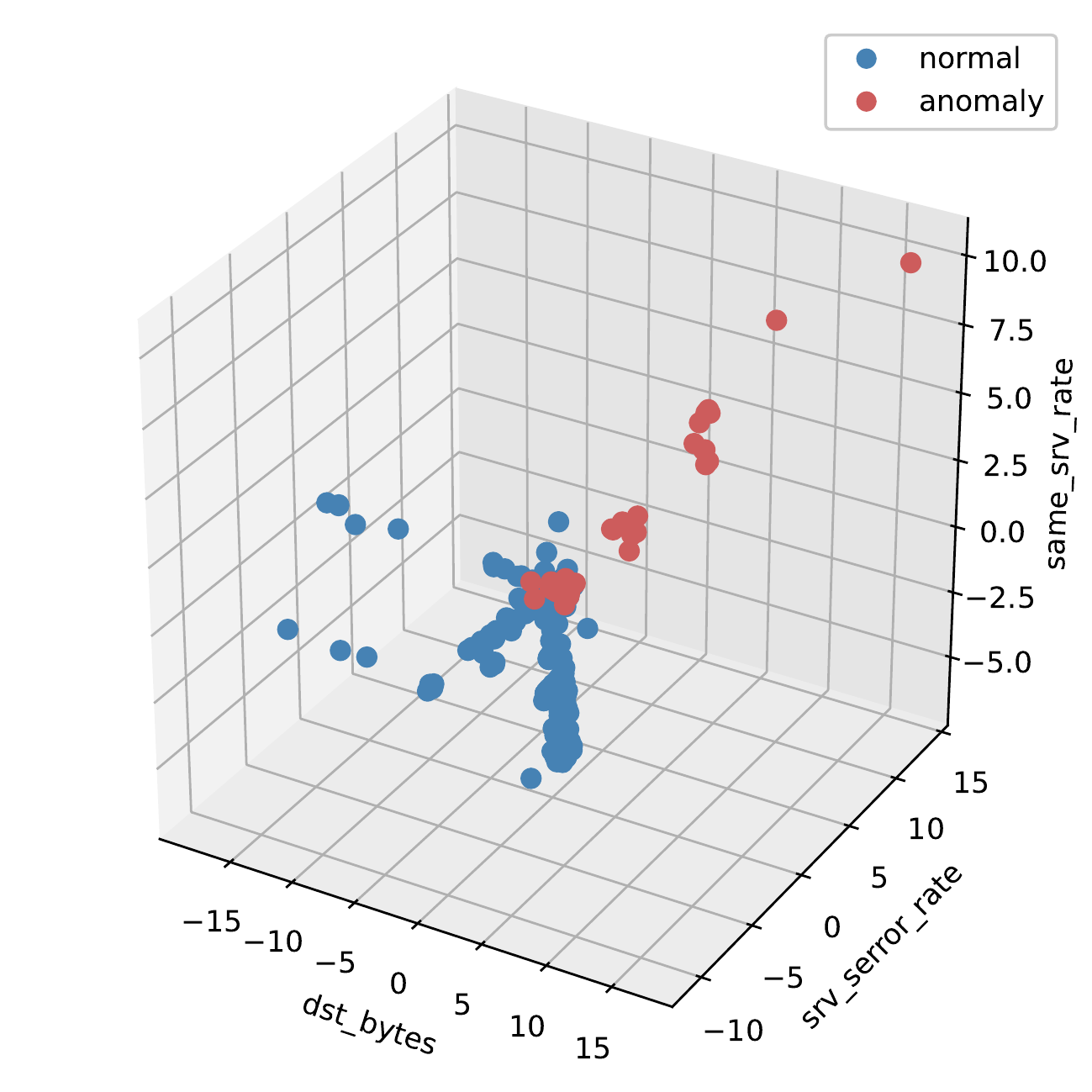}
      \caption{}
      \label{fig:sc-fed-transformed}
    \end{subfigure}
  \caption {(a) Original records. (b) Reconstructed records using Centralized PCA. (c). Reconstructed records using Fed-PCA.}
  \label{fig:visulization}
\end{figure*}


\begin{figure*}[t]
	\centering
	\begin{subfigure}{0.3\textwidth}
		\centering
		\includegraphics[width=0.9\textwidth]{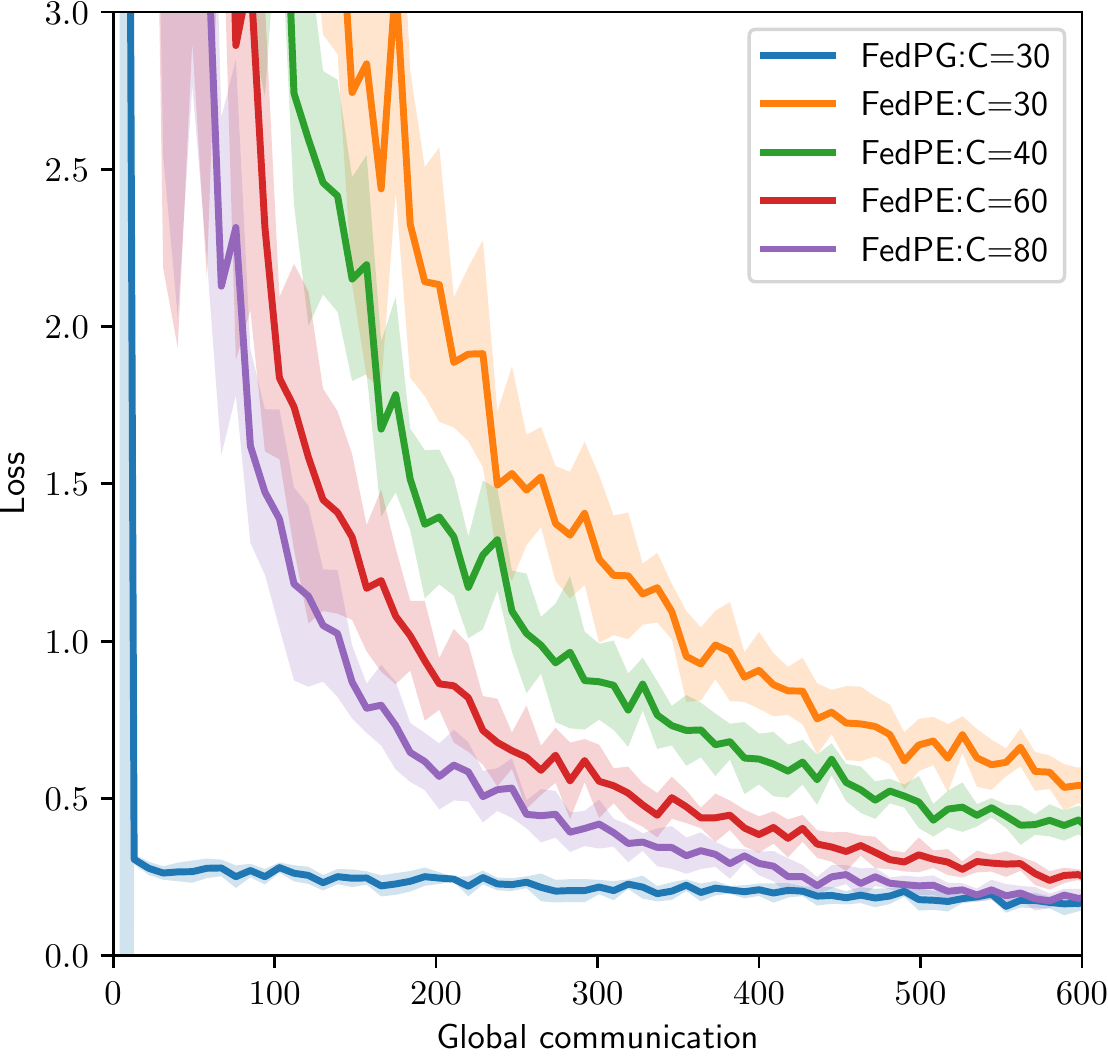}
		\caption{}
		\label{fig:loss}
	\end{subfigure}
	\begin{subfigure}{0.3\textwidth}
		\centering
		\includegraphics[width=0.9\textwidth]{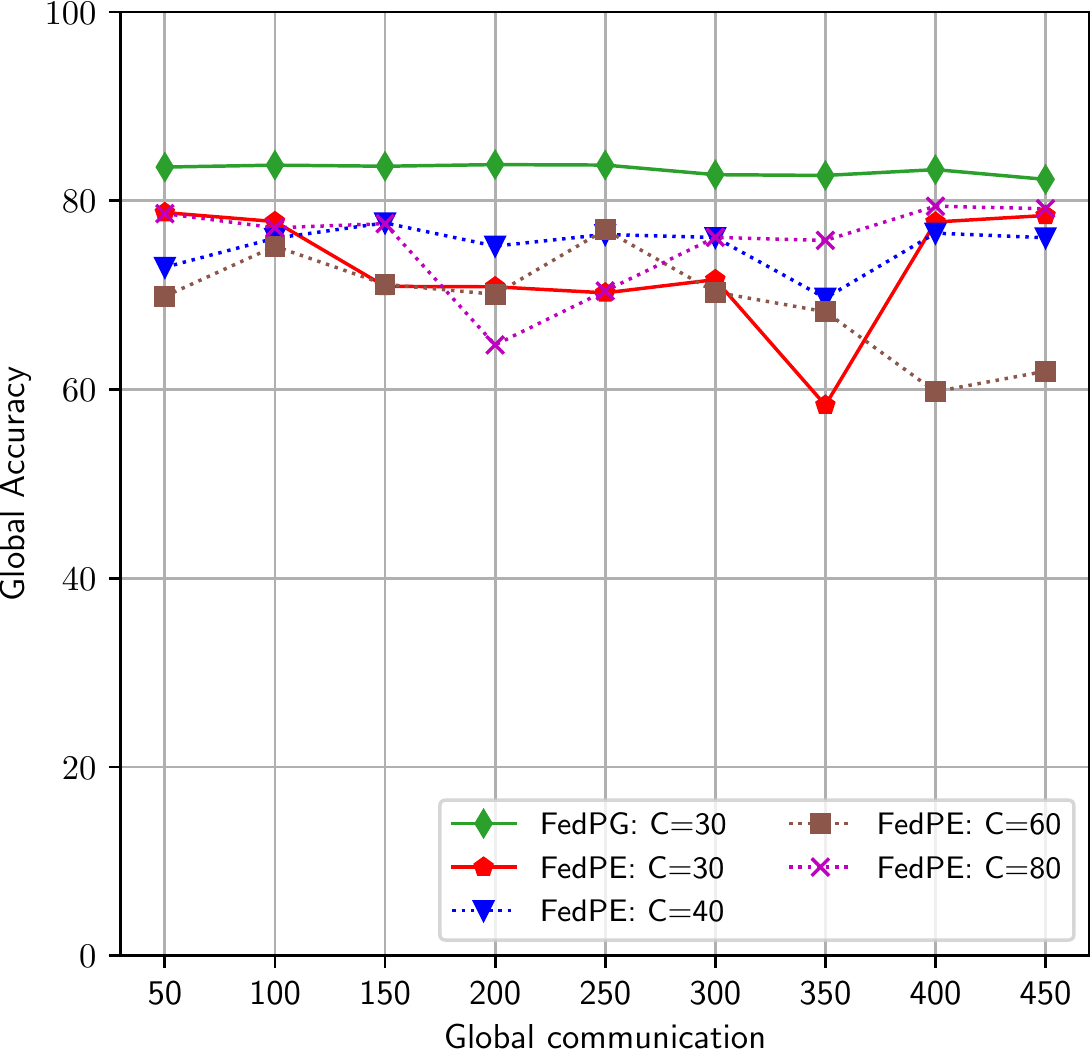}
		\caption{}
		\label{fig:acc}
	\end{subfigure}
	\begin{subfigure}{0.3\textwidth}
		\centering
		\includegraphics[width=0.9\textwidth]{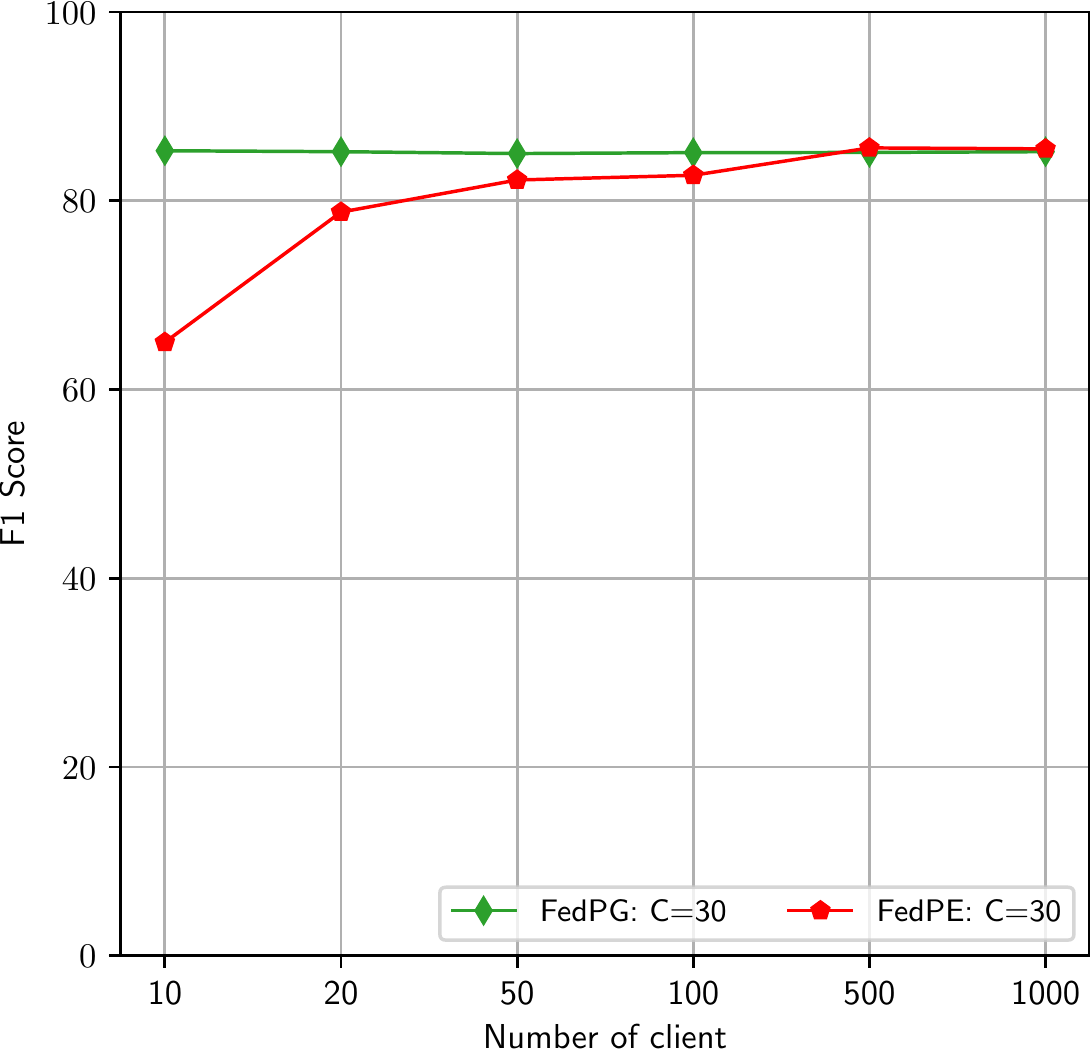}
		\caption{}
		\label{fig:F1-clients}
	\end{subfigure}
	\caption {(a) Training loss, (b) Accuracy vs. Global communication $T$; (c) Effect of number of clients $N$ on FedPE and FedPG. }
	\label{fig:convergence}
\end{figure*}

To demonstrate the capability of anomaly detection of our proposed methods, we compare the performance of Fed-PCA with the following baselines methods: Centralized PCA, Self-learning PCA, AutoEncoder~\cite{Ieracitano2020}, and Long short-term memory (LSTM)~\cite{Naseer2018}. In Centralized PCA, all records are retrieved to global server to learn a global model without considering the setting of IDS. In Self-learning PCA, we train the model of each clients using the local records only, then the average performance over all clients is reported. The performance of Fed-PCA is obtained by using FedPG.

As defined in Sec.~\ref{Sec:dataset}, the threshold $p$ for intrusion detection defines a fundamental trade-off between false positive rate (FPR) and true positive rate (TPR). The detailed performance of Fed-PCA with different thresholds is shown in Fig.~\ref{fig:results-p}. We can see that smaller thresholds lead to higher TPR but larger FPR, while larger thresholds result in lower TPR and smaller FPR. We observe that all the evaluations of Fed-PCA quickly reach over 0.8 with a low FPR ($<0.1$), which makes the detection system effective and stable. In practice, one can choose a large threshold if a low FPR is the main objective. Here, we select $p=0.5$ for all methods and compare the detection performance with baselines in Table~\ref{tab:performance}.

We observe that the proposed Fed-PCA has the best performance with an attack detection rate of $85.82\%$ F1-score. The performance of Fed-PCA is consistently close to the ones of Centralized PCA. The Self-learning PCA has worse results because it does not consider the inherent diversity of data distribution of different clients. Results show evidence that even though each client has non-i.i.d. records, the proposed Fed-PCA can still benefit from the FL algorithms to detect infusion properly without sharing the local data.

Unlike most of the DoS and Prob attacks, the R2L and U2R attacks don’t have any frequent sequential intrusion patterns as they mimic normal traffic behavior~\cite{Tavallaee2009}, which are more challenging attacks to detect, especially when there are a large number of novel sub-class attacks in the testing set as summarized in Table~\ref{tab:NLS-KDD-label}. To demonstrate the effectiveness of Fed-PCA in detecting unknown attacks, Fig.~\ref{fig:roc_U2R},~\ref{fig:roc_R2L} show two plots with ROC curves (over U2R and R2L attacks) for the unsupervised methods under different thresholds. The comparison of ROC curves shows a considerable improvement with PCA-based methods in terms of AUC score. Particularly, compared to AutoEncoder, Fed-PCA and Centralized PCA show an increase in the TPR from 70\% to 80\% at the same false alarm rate 10\% over U2R attacks. One interesting point to note is that the TPRs of Fed-PCA are lower than ones of Centralized PCA at relatively small FPRs ($<0.3$) over U2R attacks. One possible reason is that Centralized PCA have the access to the whole training set, while, due to the imbalance and scarcity of U2R attack, each client have limited access to such attacks in the federated setting. However, it is worth mentioning that for anomaly detection within an IDS development, the privacy of clients is much preferred, which can be seen as a trade-off between privacy concerns and detection performance.
\\

\subsubsection{Visualization of projections onto principal components}
To further investigate the effectiveness of the proposed method, we use a scatterplot to visualize the transformation of traffic records over U2R using Fed-PCA in Fig.~\ref{fig:visulization}. We first select three features\footnote{Network features in Fig.~\ref{fig:visulization}: number of bytes ('dst\_bytes'), rate of SYN error ('srv\_serror\_rate'), rate of same service connection ('same\_srv\_rate')} for plotting, then reconstruct the original records from the principal components learned by Fed-PCA. We can see that Fed-PCA basically transforms original records from an existing coordinate system into a new coordinate system, which is able to differentiate between normal records and anomalies. This proves that when Fed-PCA reconstructs the original records from the principal components, it will have a small reconstruction error for normal data but a large error as discussed in Sec.~\ref{Sec:PCA-for-AD}. Note that the Centralized PCA demonstrates the similar projections with different ranges of values, which does not affect the detection performance of it as shown in Table~\ref{tab:performance}.
\\
\subsubsection{Efficiency of Grassmann Manifold Algorithm}~\label{Sec:results-grassmann}
We study the convergence performance of both Fed-PCA algorithms i.e., Federated PCA over Grassmann manifold and Euclidean space, denoted as FedPG and FedPE, in Fig~\ref{fig:loss}, \ref{fig:acc}. We see that, given the same number of local communication round $C=30$, FedPG converges much faster than FedPE. Also, Table~\ref{tab:training-time} shows that FedPG requires less training time than FedPE. The latter algorithm FedPE can achieve a faster convergence rate by increasing $C$, however, leading to longer training time. We then study the sensitivity of the proposed algorithms with respect to the number of clients. We plot the detection performance of the proposed algorithms versus different numbers of clients in Fig.~\ref{fig:F1-clients}. One interesting point we observe is that the performance of FedPE increases when the number of clients increases. On the contrary, the performance of FedPG is constantly better and stable even when the number of clients is limited. This is because the learned parameters of FedPG lie on the Grassmann manifold created by the constraint~(\ref{eq:grass-constraint}), which leads to fast convergence in terms of iteration as discussed in Sec.~\ref{Sec:Complexity}. It shows evidence that FedPG achieves stable detection performance with less training time, which is much preferred in a realistic IDS development setting.

\comment{
\begin{figure*}[t]
    \centering
    \begin{subfigure}{0.3\textwidth}
      \centering
      \includegraphics[width=\textwidth]{Experimental_results/Federated_PCA_Loss_2.pdf}
  	  \caption{}
      \label{fig:loss}
    \end{subfigure}
    \begin{subfigure}{0.3\textwidth}
      \centering
      \includegraphics[width=\textwidth]{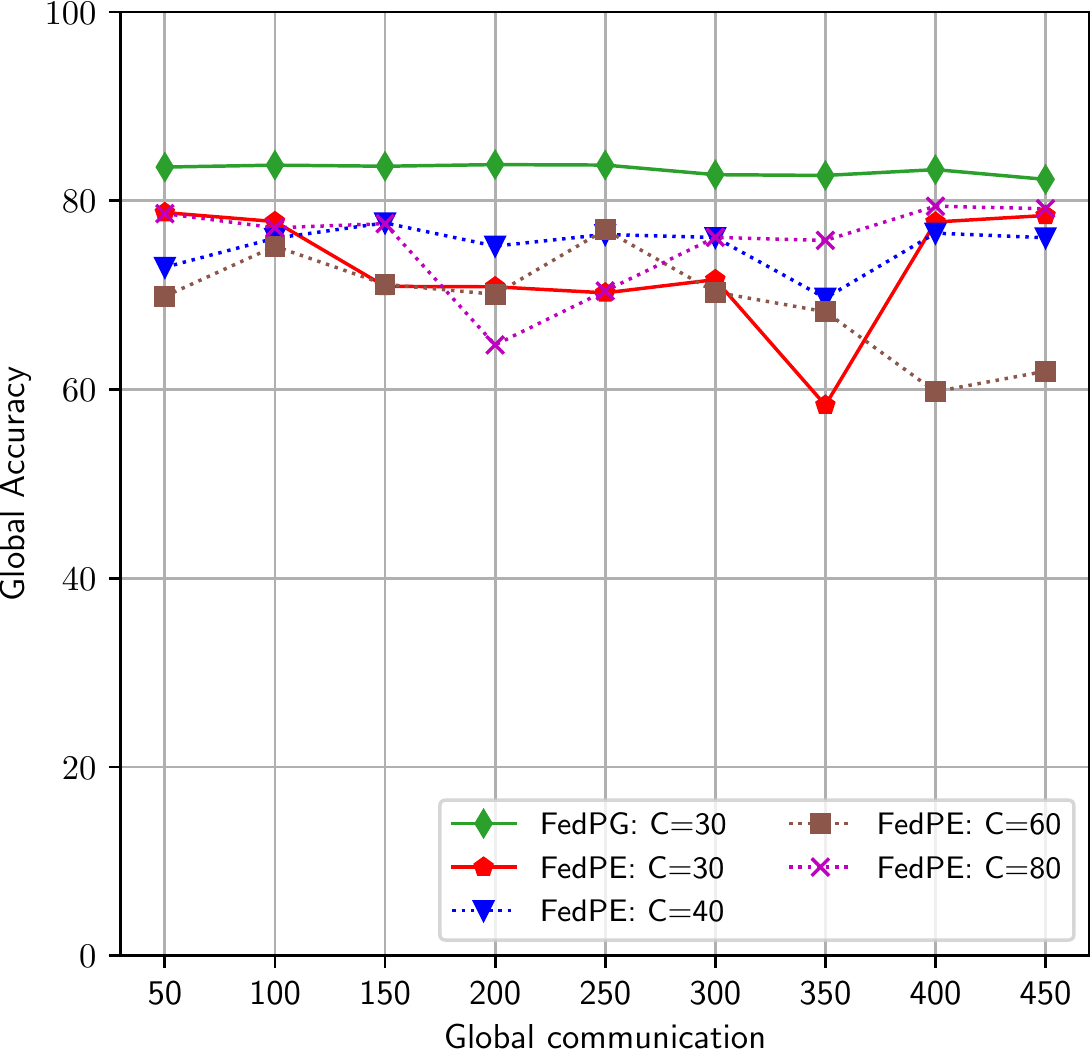}
  	  \caption{}
	  \label{fig:acc}
    \end{subfigure}
    \begin{subfigure}{0.3\textwidth}
      \centering
      \includegraphics[width=\textwidth]{Figure/F1-clients.png}
  	  \caption{}
	  \label{fig:F1-clients}
    \end{subfigure}
  \caption {(a) Original records. (b) Reconstructed records using Fed-PCA. (c). Reconstructed records using Centralized PCA. )}
  \label{fig:convergence}
\end{figure*}
}

\begin{table}[t]
	\centering
	\caption{Wall clock training time with 1000 global iterations}
	\label{tab:training-time}
	\begin{tabular}{|m{2.2 cm}|M{3cm}|M{2cm}|}
		\hline 
		\textbf{}             & \textbf{Average training time for each client (seconds)} & \textbf{Total training time (seconds)} \\ \hline  \hline
		\textbf{FedPG: C=30 }                & \textbf{0.036}                                     & \textbf{75.10 }                   \\ \hline 
		FedPE: C=30                 & 0.048                                     & 99.03                    \\ \hline
		FedPE: C=40                 & 0.064                                     & 130.52                    \\ \hline
		FedPE: C=60                 & 0.097                                     & 195.97                    \\ \hline
		FedPE: C=80                 & 0.131                                     & 262.69                    \\ \hline
	\end{tabular}
\end{table}

\section{Conclusion}

In this paper, we introduce a novel federated learning framework, namely Federated PCA on Grassmann manifold, in which IoT devices work jointly to profile benign network traffic without exchanging local data. Specially, we first design FedPE, a federated algorithm for approximating the PCA based on the idea of ADMM with theoretical guarantees. We then formulate the federated PCA via Grassmannian optimization and propose the second algorithm FedPG to ensure a fast training process and early detection in a resource-limited computing environment. Experimentally, we show that FedPE and FedPG are well-adapted to the IoT environment and demonstrate that the proposed methods outperform baseline approaches on the NSL-KDD dataset. 
\bibliographystyle{IEEEtran}
\bibliography{references} 
\end{document}